\documentclass[letterpaper, 10 pt, conference]{ieeeconf}  

\IEEEoverridecommandlockouts                              

\usepackage{graphicx}
\usepackage{subcaption}
\usepackage{rotating}
\usepackage{amsmath}
\usepackage{float}
\usepackage[usenames,dvipsnames]{xcolor}
\usepackage{tikz}
\usepackage{amsmath, amssymb, amsfonts}
\usepackage[ruled,vlined,linesnumbered]{algorithm2e}
\usepackage{array}
\usepackage{algorithmic}
\usepackage{graphbox}
\usepackage{wrapfig}
\usepackage{makecell}
\usepackage{booktabs}
\usepackage{url}
\usepackage{balance}
\usepackage{cite}

\newcommand{\zw}{\textcolor{black}}
\newcommand{\zz}{\textcolor{black}}

\title{\LARGE \bf Learning Wheelchair Tennis Navigation from Broadcast Videos with \zw{Domain Knowledge Transfer and }Diffusion Motion Planning}

\author{Zixuan Wu$^{1\dagger}$, Zulfiqar Zaidi$^{1\dagger}$, Adithya Patil$^{1\dagger}$, Qingyu Xiao$^{1}$ and Matthew Gombolay$^{1}$
\thanks{{$^\dagger$} Equal contribution}
\thanks{$^{1}$ {Authors are affiliated with the Georgia Institute of Technology, Atlanta, GA 30332 USA. Corresponding author: Zixuan Wu (zwu380@gatech.edu)}}
}

\begin{document}

\maketitle
\thispagestyle{empty}
\pagestyle{empty}

\begin{abstract}    
In this paper, we propose a novel \zw{and generalizable} \zz{zero-shot} knowledge transfer framework that distills expert sports navigation strategies \zw{from web videos} into robotic systems with adversarial constraints and out-of-distribution image trajectories. Our pipeline enables diffusion-based imitation learning by reconstructing the full 3D task space from multiple partial views, warping it into 2D image space, closing the planning loop within this 2D space, \zw{and transfer constrained motion of interest back to task space}. Additionally, we demonstrate that the learned \zz{policy}
can serve as a local planner in conjunction with position control. \zw{We apply this framework in the wheelchair tennis navigation problem to} guide the wheelchair into the ball-hitting region. Our pipeline achieves a navigation success rate of 97.67\% in reaching real-world recorded tennis ball trajectories with a physical robot wheelchair, and achieve a success rate of 68.49\% in a real-world, real-time experiment on a full-sized tennis court\footnote[2]{\ Code is at \url{https://github.gatech.edu/MCG-Lab/tennis_gameplay_learning}}. 
\end{abstract}
\section{INTRODUCTION}

\zz{
Leveraging web videos of expert performance offers a promising way to train robots, particularly in agile robotics and sports, where gathering expert demonstrations—such as in wheelchair tennis—is extremely challenging. Numerous human-designed mechanical prototypes have demonstrated capabilities in various sports, including table tennis \cite{Buchler2022learning,lee_ping_colab_2023, d2024achieving}, badminton \cite{Mori2019highspeed, liu_badminton_2013}, tennis \cite{zaidi2023athletic, MRMC_2024, yang2022varsm, krishna2022utilizing}, and soccer \cite{haarnoja2024learning}. Although humans and robots share similar decision loops (e.g., perception, planning, and control)~\cite{siegel2020robotics}, research in robotic sports has yet to explore learning agile and dynamic behaviors from web expert videos, a gap we aim to address. Prior work in robotic sports has primarily focused on ball motion prediction~\cite{nakashima2014online, zhang2010visual} and motion planning~\cite{kocc2016new, zhang2013motion} using techniques ranging from sophisticated hybrid ball dynamics models~\cite{kocc2016new, serra2016optimal} to simplified approaches like virtual hitting points~\cite{mulling2011biomimetic} or linear collision models~\cite{ji2021model}. Reinforcement learning (RL)~\cite{josef2020deep} and imitation learning (IL)~\cite{wen2023any} have also shown promise in sports robotics~\cite{gao2020robotic, abeyruwan2023sim2real}, but RL requires extensive training in simulated environments and IL depends on costly expert demonstrations.
}

Images and videos have become popular sources of demonstrations for IL \cite{chi2023diffusionpolicy, zhao2023learning}.
Although advanced generative models can reason within image space, their imitation policies typically rely on static objects without autonomous dynamics (e.g., manipulated objects in pick-and-place tasks) \cite{chi2023diffusionpolicy, bahl2023affordances, chen2021learning}.
Additionally, these methods often require \zz{costly} human demonstrations captured with depth or multi-view cameras. 
To overcome the limited availability of learning sources, research has suggested using web videos from monocular cameras as a vast, accessible resource.
However, these videos are often noisy and incomplete, necessitating information extraction to obtain valuable data. Notably, Zhang at al. \cite{zhang2023vid2player3d} demonstrated that a simulated agent could learn to play tennis using data collected from broadcast videos using a combination of IL and RL, but this approach was only applied in simulation.
\begin{figure}
    \centering
    \includegraphics[width=0.8\linewidth]{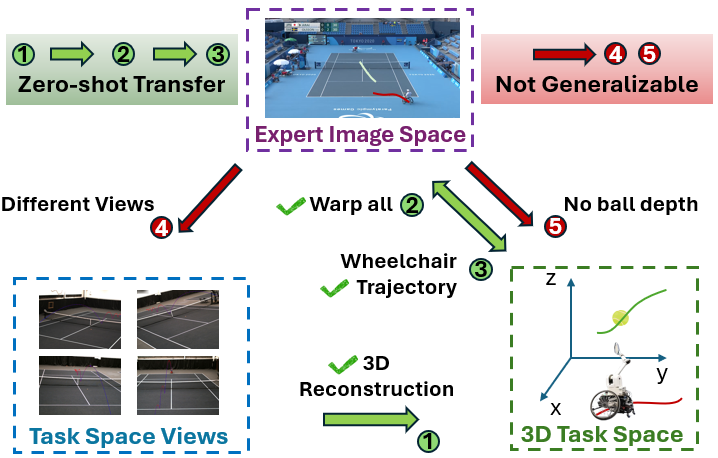}
    \caption{It is not possible to transfer knowledge directly from the expert videos to our task spaces due to the lack of ball depth in the expert image space and different camera angles to the task space (red arrows). As such, we propose a three-step approach (green arrows): 1) reconstruct full 3D motion in the task space (ball-wheelchair), 2) project all motions into the 2D to apply the imitation policy, and 3) transfer planar wheelchair motion back to the 3D task space.}
    \label{fig:motivation}
    \vspace{-5mm}
\end{figure}
Our work 
differs from previous video learning approaches \cite{chi2023diffusionpolicy, zhao2023learning, torabi2018behavioral, chang2020semantic} \zw{in two ways.} First, 
\zz{we utilize an autonomous wheelchair robot that operates in a challenging, adversarial setting, with less than two seconds to intercept an incoming tennis ball moving under its own dynamics.}
This requires our IL model to respond to dynamic \zz{ball} trajectories in real-time, rather than relying on static text commands or navigation goals. \zw{Second, the disparity in perspectives between web videos and our real-world setting makes 2D-to-2D generalization impossible, and the lack of tennis ball depth information in monocular video streams prevents 2D-to-3D scene reconstruction.} 
\zz{
To address this out-of-distribution (OOD) challenge, we reconstruct the 3D scene from multiple partial camera views in our task space, project the objects into the expert's 2D image space, run the imitation policy there, and transfer the resulting 2D robot motion back to the 3D task space (see Figure \ref{fig:motivation}).
}




Our contributions are as follows:

\begin{itemize}
\item To best of our knowledge, we are the first to tackle the web video learning problem in agile navigation under adversarial settings with OOD \zz{task space}. 
\zw{We propose a novel} \zz{zero-shot} \zw{knowledge transferring framework with a diffusion motion planner that} can also be generalized to other 2D mobile robot navigation problems.
\item \zw{We incorporate mean-shift matchplay identification and audio segmentation model into our online broadcast wheelchair tennis data extraction pipeline, which is able to handle raw, unprocessed broadcast footage and help efficiently produce training-ready data.}
\item We design a real-time feedback \zw{planning-control} system, including a PD controller guided by our IL policy, and demonstrate zero-shot transfer to a real robot on a real tennis court. \zw{We show our method outperforms baselines in multi-level ablation studies.}


\end{itemize}

\section{RELATED WORK}
\label{sec:related_works}
\subsection{Robotics in Athletics}


Robotic platforms in athletics can be classified by maneuverability into fixed, locally moving, and globally moving systems. Fixed platforms, like stationary manipulators for table tennis \cite{Buchler2022learning} and badminton \cite{Mori2019highspeed}, are limited to returning balls within a small range. Locally moving platforms, enhanced with gantry rails \cite{d2023robotic, MRMC_2024, liu_badminton_2013}, offer more mobility, but remain restricted in reach and are often costly \cite{MRMC_2024}.

This paper focuses on a global court navigation system for a wheelchair tennis robot \cite{zaidi2023athletic}, using a mobile electric wheelchair base. 
While similar to previous systems \cite{dong2020catch, yang2022varsm} due to its mobile base, our platform differs in that the wheelchair robot meets Paralympic tennis standards.

\subsection{Vision-based Planning and Control}
Visual information has long been used in robot planning and control, starting with visual servo techniques \cite{chaumette2006visual, chaumette2007visual} and visual teach-and-repeat methods.
Recent works use visual prompts \cite{gu2023rt}, affordances \cite{bahl2023affordances}, or vision-based rewards \cite{chen2021learning} to inform motion planning via machine learning on videos. For example, RT-Trajectory \cite{gu2023rt} generates actions based on static sketched, demonstrated, or language-guided trajectories. 
In contrast, our approach generates trajectories using a diffusion model that responds to \zz{dynamic} tennis ball trajectories.
Although reward functions can be learned from videos to score optimal trajectories \cite{chen2021learning}, these methods rely on separate visual model predictive control (VMPC) for trajectory sampling, limiting their ability to perform \zz{zero-shot} motion planning.
Bahl et al. \cite{bahl2023affordances} address this by generating trajectories using a Transformer with learned visual affordances, but their approach operates slowly in static environments without feedback, making it unsuitable for our fast-paced, adversarial game setting.



Few studies shift the decision or planning loop into the image space, as most prior work generates \cite{mezouar2002path, feng2022image} or optimizes \cite{keshmiri2016image} trajectories in 
\zz{3D} task space before projecting them into the image space.
In contrast, we aim to learn a navigation policy directly from web videos, eliminating the need for hand-crafted motion planning, costly wheelchair player demonstrations or environment interactions.
While prior work has used video features for reward shaping \cite{peng2018sfv, xiong2021learning} or goal conditions \cite{bahl2023affordances} in RL, these methods still depend on human demonstrations and environment interactions.

\subsection{Motion Planning with Imitation Learning}

IL based motion planning originated with behavior cloning (BC) and its application in autonomous driving, using images and laser inputs to predict direction \cite{pomerleau1988alvinn}. The issue of compounding errors in BC over time \cite{ross2010efficient} was addressed by the Dataset Aggregation (DAgger) algorithm \cite{ross2011reduction}, which requires interaction with expert policies and the environment, contrary to the \zz{zero-shot} 
learning approach we envision. 
The pseudo-action labeling approach, such as BCO \cite{torabi2018behavioral}, learns a dynamic model for mobile robots during exploration  to estimate inter-frame actions in observations and construct experience tuples for BC. Although this algorithm has been applied to visual navigation problems by associating YouTube video frames with actions and rewards in Q-learning \cite{chang2020semantic}, it requires ego-centric observations, pre-defined navigation goals, and real-world exploration by the robot, which are impractical in our context.


Diffusion probabilistic models are generative models that denoise Gaussian noise into in-distribution samples under certain constraints \cite{sohl2015deep, ho2020denoising, perez2018film, rombach2022high}. To tackle the state drift problem in Markov-based BC, researchers have proposed diffusion-based long-horizon path planning, demonstrating both global effectiveness and local consistency \cite{janner2022planning, wu2024diffusionreinforcement, sean2023diffusion}. However, these global plans are generated only once with initial constraints, lacking in-task feedback and real-time performance guarantees. Recently, diffusion \cite{chi2023diffusionpolicy} and transformer-based trackers \cite{wen2023any, zhao2023learning} have been proposed to mitigate compounding errors by integrating sequential position control
or guiding policy learning. However, they still rely on static objects or text inputs and sometimes require action labels \cite{wen2023any}.
\section{Methodology}

\begin{figure}[t]
    \centering    
    \includegraphics[width=\linewidth]{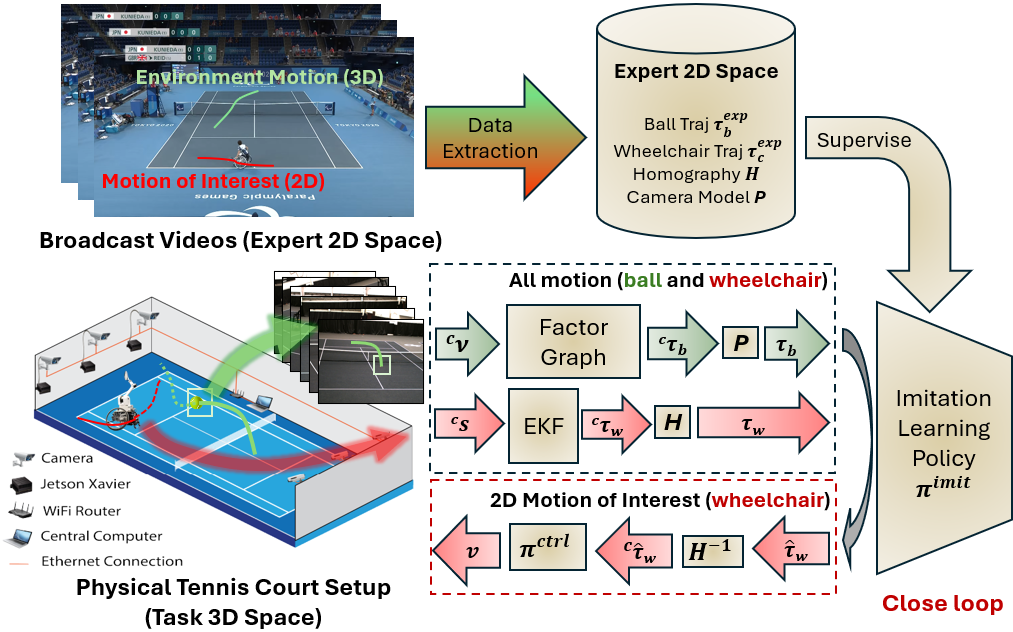}
    \caption{Overview of our \zw{knowledge transferring} framework: \zw{The wheelchair motion is the motion of interest (MOI, indicated by red) since we need to control it in our task space. The ball motion is the environment motion (Env-M, indicated by green) on which the wheelchair should condition. The knowledge transfer is valid with the only assumption that MOI is 2D motion and Env-M can be 3D.} }
    \label{fig:system_overview}
    \vspace{-6mm}
\end{figure}

\zz{In this work, we develop a pipeline that extracts data from wheelchair tennis broadcast videos, learns a policy on the extracted data in image space, and transfers it to a real-world wheelchair robot. 
}
Our dataset and implementation scenarios span two distinct domains: the training data is derived from monocular 2D broadcast video footage of the 2020 Paralympics tennis game, while the algorithm must be applied in a real 3D tennis court environment with different views from external wall cameras (see Figure \ref{fig:motivation}-\ref{fig:system_overview}). This discrepancy poses significant challenges for knowledge transfer. Additionally, we aim to achieve \zz{zero-shot} sim-to-real transfer without any interactive fine-tuning.

\zz{
Our method addresses these challenges through four components: (1) data extraction from broadcast videos (Sec.~\ref{subsec:data_extraction}), (2) a knowledge transfer framework (Sec.~\ref{subsec:imitation_learning_framework}), (3) imitation policy learning (Sec.~\ref{subsec:imitation_policy_learning}), and (4) real-time feedback control for execution on the robot (Sec.~\ref{subsec:feedback_control}).
}

\subsection{Data Extraction from Broadcast Videos}
\label{subsec:data_extraction}

\begin{figure}[t!]
    \centering    
    \includegraphics[height=0.6\linewidth]{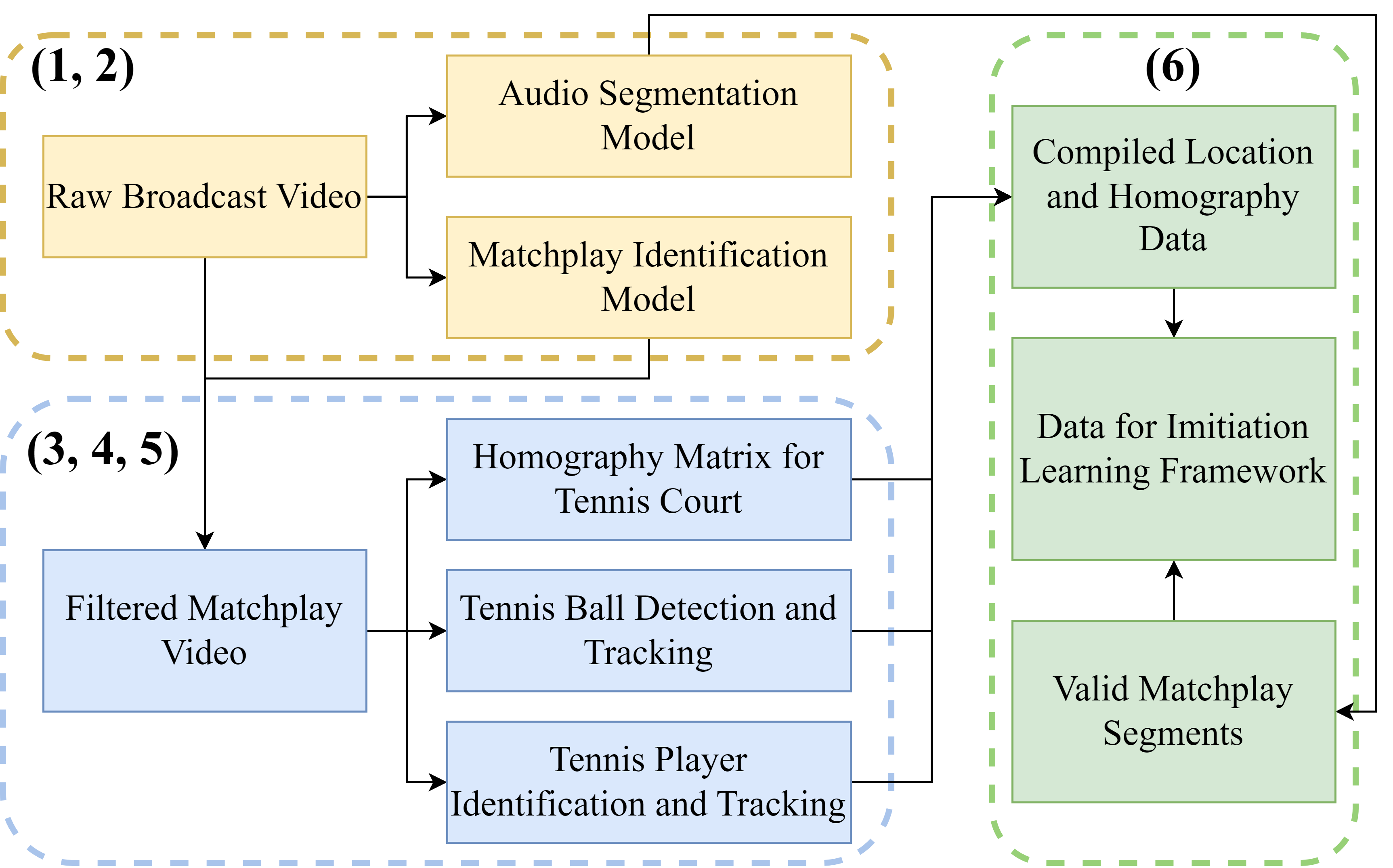}
    \caption{Various steps involved in creating the dataset from broadcast footage of Paralympic 2020 games.}
    \label{fig:dataset_creation}
    \vspace{-5mm}
\end{figure}

\zz{
The goal of this phase is to extract player movement data from broadcast wheelchair tennis videos. We used footage from the singles matches from the Tokyo 2020 Paralympic Games. The extraction process involves six steps (Fig.~\ref{fig:dataset_creation}): (1) Using a mean-shift model to filter non-matchplay segments, (2) Training a neural network to detect ball hit audio, (3) Obtaining court homography, (4) Tracking the tennis ball, (5) Locating the player on the nearside of the court, and (6) Aggregating the data for training output.
}


\zz{
First, we train two models: one to filter non-matchplay footage and another to detect ball hits using match audio. The matchplay filtering model applies Mean Shift clustering with a flat kernel to preprocessed video frames---resized, Gaussian filtered, converted to grayscale, and intensity rescaled---to retain only matchplay. The algorithm updates each data point, $x_i$ (i.e., a frame), using \( x_i \leftarrow \frac{\sum_j K(x_j - x_i) x_j}{\sum_j K(x_j - x_i)} \), where \( K \) is a flat kernel. We found that mean shift clustering is a more stable and robust method for filtering than approaches based on court line detection. The hit detection model utilizes a fully connected neural network with three layers (sizes 256, 128, and 2) on audio features extracted via Mel-frequency cepstral coefficients (MFCCs), chromagrams, and mel-spectrograms~\cite{muller2015fundamentals}. The network is trained to minimize the cross-entropy loss, \( L = -\left[ y \log(\hat{y}) + (1 - y) \log(1 - \hat{y}) \right] \), where $y$ is a binary indicator for hit, on manually labeled audio clips of hits and randomly selected non-hit clips. Prior methods~\cite{huang2019tracknet,Bataille2022} lack this automatic filtering and hit detection capability.
}

\zz{In the third and fourth steps, we track the tennis ball using TrackNet \cite{huang2019tracknet,Bataille2022} and generate the homography matrix for court identification using the methods from \cite{Mora2018,Caspi2021}. The homography matrix translates pixel positions from the 2D image plane to 3D court coordinates, accounting for perspective distortion. The relationship between image coordinates, \((u, v)\), and court coordinates, \((x, y)\), is defined by \(\begin{bmatrix} u' & v' & w \end{bmatrix}^\top = \mathbf{H} \begin{bmatrix} x & y & 1 \end{bmatrix}^\top\), where \(\mathbf{H}\) is the homography matrix. We determine the homography matrix, \(\mathbf{H}\), by matching known court keypoints with their corresponding pixel coordinates in the image. Unlike prior methods \cite{Mora2018,Caspi2021}, we recalculate the matrix only when camera motion is detected, improving efficiency. In the fifth step, we use an open-source object detection model \cite{Jocher_Ultralytics_YOLO_2023} to detect the player on near-side of the court every ten frames, with background subtraction and contour detection combined with an Extended Kalman Filter (EKF)~\cite{moore2016generalized} for tracking in between frames. The homography matrix is then applied to determine the player's position on the court relative to the baseline.
}



\zz{
The final step involves segmenting the data into \zw{matchplay episodes (910 in total). Each episode} begins when the far-side player hits the ball and ends when the near-side player returns it, as detected using the ball hit model. \zw{Afterwards, we can examine individual episodes and optionally discard invalid ones 
(we call it semi-automatic mode if we have this manual check). }The data includes the 2D ball trajectory, the near-side player's trajectory in 2D image and 3D task space, along with the homography matrix for each frame.
}

\subsection{\zw{Knowledge Transferring} Framework Design}
\label{subsec:imitation_learning_framework}
\SetCommentSty{mybluecomment}
\newcommand{\mybluecomment}[1]{\textcolor{gray}{\small\textit{#1}}}
\SetKwComment{Comment}{$\lhd$ }{}
\begin{algorithm}[h]
\SetAlgoLined
\KwData{$\mathcal{D}$ (expert videos), $^{c}\mathcal{T}_b, ^{c}\mathcal{T}_w$ (tennis ball and wheelchair trajectory in task space), $^{c}X_w$ (wheelchair position on the court in task space), $^{c}\mathcal{V}$ (multiple partial views from wall cameras on the court in task space), $^{c}\mathcal{S}$ (wheelchair on board sensors)}
\Comment{Extract ball and wheelchair trajectories $\mathcal{T}_b^{\text{exp}}, \mathcal{T}_w^{\text{exp}}$, camera model $P$ and homography $H$ from expert video} 
$\mathcal{T}_b^{\text{exp}}, \mathcal{T}_w^{\text{exp}}, P, H = \text{VideoTracker}(\mathcal{D})$\; \label{eq:extract}
\Comment{Learn the imitation policy $\pi^{\text{imit}}$} 
$\pi^{\text{imit}} = \text{TrainPolicy}(\mathcal{T}_b^{\text{exp}}, \mathcal{T}_w^{\text{exp}})$\; \label{eq:imitation}
\Comment{Reconstruct 3D trajectories with factor graph and EKF}
$^{c}\mathcal{T}_b = \text{FactorGraph}(^{c}\mathcal{V}), ^{c}\mathcal{T}_w = \text{EKF}(^{c}\mathcal{S})$\;  \label{eq:fg}
\Comment{Warp trajectories $^{c}\mathcal{T}_b, ^{c}\mathcal{T}_w$ into the expert image space}
$\mathcal{T}_b = P(^{c}\mathcal{T}_b), \mathcal{T}_w = H(^{c}\mathcal{T}_w)$\;  \label{eq:warp}
\Comment{Apply the imitation policy on the expert image space}
$\hat{\mathcal{T}}_w = \pi^{\text{imit}}(\mathcal{T}_b, \mathcal{T}_w)$\; \label{eq:imit}
\Comment{Convert desired wheelchair trajectory to task space $^{c}\hat{\mathcal{T}}_w$}
$^{c}\hat{\mathcal{T}}_w = H^{-1}(\hat{\mathcal{T}}_w)$\; \label{eq:totask}
\Comment{Execute the control policy $\pi^{\text{ctrl}}$ on the robot}
$v = \pi^{\text{ctrl}}(^{c}X_w, ^{c}\hat{\mathcal{T}}_w)$\label{eq:feedback}
\caption{\zw{Knowledge Transferring} Framework}
\label{alg:pseudocode_imitation_learning_framework}
\end{algorithm}


We utilize \zz{our} web-sourced Paralympics' wheelchair tennis videos (\zw{expert 2D image space}) as our training dataset (Sec.~\ref{subsec:data_extraction}), with the objective of executing the learned policy on a robot in a physical tennis court (\zw{3D} task space). Given that our expert video data comes from a single camera angle without depth (similar to TV broadcasts) different from \zw{task space} view configurations, it is neither feasible to fully reconstruct \zz{expert data}
in the \zz{3D} task space nor perform direct 2D-2D knowledge transfer. To address this, we design an algorithm (Algorithm~\ref{alg:pseudocode_imitation_learning_framework}) that bypasses the need for full 3D scene reconstruction by leveraging \zw{motion and} visual geometry constraints through the following steps (subscript \textit{w} means wheelchair and \textit{b} means the tennis ball): (1) extract training data from broadcast videos (line \ref{eq:extract}); (2) train the imitation policy in the 2D image space using the extracted data (line \ref{eq:imitation}); (3) \zz{in a real-world court setup,} reconstruct the ball and wheelchair trajectories \zz{in 3D task space} using multiple partial views from wall cameras and  onboard wheelchair sensors, leveraging a factor graph~\cite{xiao2024multi} and \zw{EKF} (line \ref{eq:fg}); (4) 
project the 3D locations of the ball and wheelchair into the 2D image space using the camera model (line \ref{eq:warp}); (5) run the imitation policy to generate the desired wheelchair trajectory in the 2D image space (line \ref{eq:imit}); (6) transform the desired wheelchair trajectory from the 2D image space to the task space using the inverse homography matrix (line \ref{eq:totask} - possible since the wheelchair always remains on the ground plane); and (7) execute the trajectory on the wheelchair in real-time using a feedback controller (line \ref{eq:feedback}). \zw{The process is visualized in Fig.~\ref{fig:system_overview}, and our knowledge transferring framework can be generalized to other 2D constrained motion settings, multi-agent collaborations and imitation policy structures (Fig. \ref{fig:generalization}).}

\subsection{Imitation Policy Learning}
\label{subsec:imitation_policy_learning}


\zw{
Our imitation policy operates in the image space, where noise and uncertainty affect web video data and task space trajectory estimation. To address this, we employ a diffusion policy that recovers in-distribution trajectories from noisy inputs, with prior work showing strong generalizability with corrupted data \cite{chen2024slightcorruptionpretrainingdata} and image-space operations \cite{wen2023any} (see Algorithm \ref{alg:diffusion_pseudocode}).}
First, we segment the wheelchair and ball trajectories to an appropriate length of 32 and 18 timesteps for the input and output respectively
(line \ref{eq:cut_traj}). A batch of random trajectories $\tau_0$ is drawn from the dataset (line \ref{eq:draw_traj}) and Gaussian noise is applied using a \zz{cosine schedule} (line \ref{eq:schedule_noise}). The training step involves calculating the MSE between the applied noise and the noise predicted by our model $s_{\theta}$ and updating the model via gradient descent (line \ref{eq:gradient_descend}). During sampling, we apply our denoising model $s_{\theta}$ to random Gaussian noise and enforce constraints to ensure that the diffusion trajectory starts at the current location (line \ref{eq:sample}).

Additionally, we train other baseline policies using our framework: the Action Chunking with Transformers (ACT) model \cite{zhao2023learning}, a fully connected regression neural network (FCR), and an autoencoder combined with a fully connected regression neural network (AE+FCR). In the AE+FCR model, the autoencoder encodes the prior ball trajectory, and the fully connected network predicts the commanded trajectory based on the wheelchair's current position and the encoded prior ball trajectory.

\begin{algorithm}[t] 
\SetAlgoLined
\KwData{$\mathcal{T}_w, \mathcal{T}_b(\text{wheelchair and ball image trajectories})$} 
\Comment{Extract the trajectory pieces $\mathcal{T}_b^{sub}, \mathcal{T}_w^{sub}$}
$\mathcal{T}^{sub} = \mathcal{T}_b^{sub}, \mathcal{T}_w^{sub} = Segment(\mathcal{T}_b, \mathcal{T}_w) \label{eq:cut_traj}$\;
\While{Training iteration}
{ 
\Comment{Draw training trajectory $\tau$, diffusion timestep $i$, and random noise $\epsilon$}
$\tau_0, m \sim \mathcal{T}^{sub}; i \sim \mathcal{U}(1, T), \epsilon \sim \mathcal{N}(0, I);$
\label{eq:draw_traj} \\
\Comment{Apply the scheduled noise to the drawn trajectory}
$\tau_i \leftarrow \sqrt{\bar{\alpha}_i} \tau_0 + \sqrt{1 - \bar{\alpha}}_i \epsilon$\;\label{eq:schedule_noise}
\Comment{Compute MSE Loss and update model $s_{\theta}$}
$\mathcal{L}(\theta) = \| \epsilon - s_\theta(\tau_i, i, m))\|_2^2;\ \theta = \theta + \alpha \nabla_\theta \mathcal{L}(\theta)$\; \label{eq:gradient_descend}
}
$\tau^T \sim \mathcal{N}(0, I)$\Comment*[r]{Sample from Gaussian noise} \label{eq:noise}
\For{all $i$ from $T$ to $1$}{ \label{line:for}
\Comment{Denoise the trajectory for one step}
$(\mu^i, \Sigma^i) \leftarrow s_{\theta}(\tau^{i}), \Sigma_{\theta}(\tau^{i}) \tau^{i-1} \sim \mathcal{N}(\mu^i, \Sigma^i)$\; \label{eq:denoise}
\Comment{Sample from denoised distribution, apply constraints}
$\tau^{i-1} \sim \mathcal{N}(\mu^i, \Sigma^i);\ \tau^{i-1} \gets \mathcal{C}(\tau^{i-1})$\label{eq:sample}
 }
\caption{Diffusion Training and Sampling}
\vspace{-1mm}
\label{alg:diffusion_pseudocode}
\end{algorithm}

\subsection{Real-time Reactive Feedback Control}
\label{subsec:feedback_control}

Once the trajectory is provided by the imitation policy in the image space, it is converted to task space using inverse homography projection and executed on the robot in real-time using a proportional-derivative (PD) position controller. Assuming the wheelchair has a heading angle $\theta$, with its current position as the origin and the desired position given by $[x, y]$, our PD controller is defined as: 
\begin{eqnarray}
    v &=& k_{1p}(x\cos\theta + y\sin\theta) + k_{1d}\dot{d} \\
    \omega &=& k_{2p}\left(\arctan2(y, x) - \theta\right) + k_{2d}\dot{\theta}
\end{eqnarray}
Here, the driving speed $v$ is proportional to the Cartesian distance error, while the angular velocity $\omega$ is proportional to the angular error. The derivative terms $\dot{d}$ and $\dot{\theta}$ help stabilize the control. Together, $v$ and $\omega$ form the commanded twist, which is converted to left and right wheel velocities by a differential drive controller and executed on the robot.




\section{Experimental Results and Discussion}
\label{sec:result}

In this section, \zz{we first evaluate the efficiency of our data extraction pipeline (Sec.~\ref{subsec:data_extraction_results}). Next,} we assess the performance of our knowledge transferring framework with diffusion imitation policy and compare it against various baselines by measuring the predictions against the expert ground-truth data (Sec.~\ref{subsec:trajectory_prediction_results}). Subsequently, we evaluate the effectiveness of our method in intercepting real-world ball trajectories \zw{in a hybrid testing (Sec.~\ref{subsec:task_space_evaluation}) and real-time real-court experiments (Sec.~\ref{subsec:real_world}).}

\begin{figure}[ht]
    \vspace{-3mm}
    \centering
    \includegraphics[width=1\linewidth]{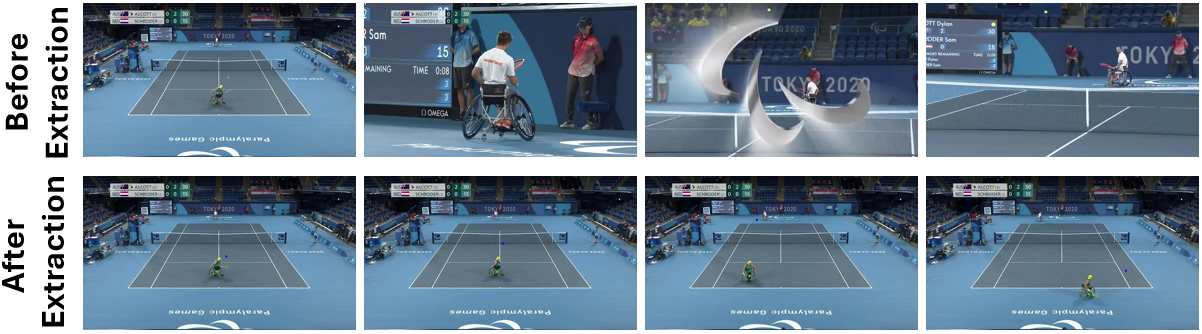}
    \caption{\zw{The pipeline filters out rest and replay frames (top) while extracting relevant gameplay data (bottom).}}
    \label{fig:extraction_pipeline}
    \vspace{-5mm}
\end{figure}

\zz{
\subsection{Data Extraction Pipeline}
\label{subsec:data_extraction_results}
\zw{Previous tennis video tracking packages \cite{huang2019tracknet,Bataille2022}  focus solely on trajectory and court line extraction. In contrast, }our data extraction pipeline automatically isolates gameplay sections and filters out irrelevant content, such as crowd shots and player close-ups (see Fig.~\ref{fig:extraction_pipeline}), removing approximately 54\% of frames from a match video and \zw{generating episodes for downstream learning tasks.} Furthermore, 
we perform player detection (using an object detection model) every 10 frames, reducing reliance on image recognition models to just 10\% of the time, further optimizing efficiency.
}

\begin{table*}[t!]
\begin{minipage}[b]{0.7\linewidth}
\centering
\caption{Benchmarking results for imitation learning models across various metrics}
\resizebox{\textwidth}{!}{
\begin{tabular}{lllcccccccc}
\toprule
 &  &  & \multicolumn{4}{c}{Semi-Automatic} & \multicolumn{4}{c}{Automatic} \\ \cmidrule(lr){4-7} \cmidrule(lr){8-11} 
\textbf{Method} & \textbf{Condition} & \textbf{Action} & \textbf{RMSE} & \textbf{DTW} & \textbf{ICP} & \textbf{Jerk} & \textbf{RMSE} & \textbf{DTW} & \textbf{ICP} & \textbf{Jerk} \\ \midrule
Diffusion (ours) & Pre 2D & $I_{\text{space}}$ & 0.321 & 6.891 & 0.206 & 1.951 & 0.359 & 7.581 & 0.259 & 4.760 \\ \midrule
Diffusion (ours) & Post 2D & $I_{\text{space}}$ & \textbf{0.277} & \textbf{5.880} & \textbf{0.167} & \textbf{1.150} & 0.346 & 7.586 & {\textbf{0.218}} & 4.360 \\ \midrule
ACT & Pre 2D & $I_{\text{space}}$ & 0.372 & 8.400 & 0.212 & 5.879 & 0.403 & 9.090 & 0.266 & 6.037 \\ \midrule
ACT & Post 2D & $I_{\text{space}}$ & 0.381 & 8.640 & 0.226 & 5.274 & 0.387 & 8.810 & 0.242 & 4.976 \\ \midrule
{FCR} & Pre 2D & $I_{\text{space}}$ & 0.345 & 7.520 & 0.216 & 5.596 & {\textbf{0.321}} & {\textbf{7.250}} & 0.236 & 1.024 \\ \midrule
{FCR} & Post 2D & $I_{\text{space}}$ & 0.287 & 6.180 & 0.183 & 4.598 & 0.399 & 9.270 & 0.243 & {\textbf{0.876}} \\ \midrule
\makecell[l]{AE+{FCR}} & Pre 2D & $I_{\text{space}}$ & 0.573 & 12.28 & 0.375 & 5.248 & 0.435 & 9.402 & 0.298 & 8.506 \\ \midrule

{Diffusion} & {Post 2D} & {$T_{\text{space}}$} & {0.716} & {15.854} & {0.552} & {4.870} & {0.747} & {16.42} & {0.586} & {5.867} \\
\bottomrule
\end{tabular}%
}
\label{tab:imitation_res}
\end{minipage}\hfill
\begin{minipage}[b]{0.29\linewidth}
\centering
\includegraphics[width=1.7in]{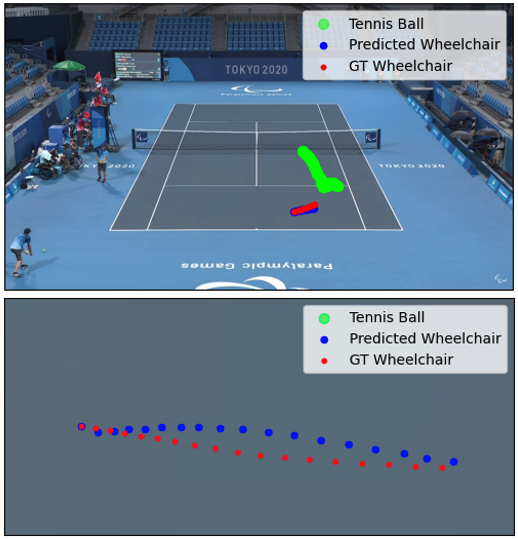}
\captionof{figure}{Diffusion Trajectory}
\label{fig:diff_path}
\end{minipage}
\end{table*}

\subsection{Trajectory Prediction from Imitation Learning}
\label{subsec:trajectory_prediction_results}

In this section, we benchmark various policies trained using our framework by comparing the predicted trajectories of each method against the ground truth wheelchair trajectories extracted from broadcast videos. The methods include the diffusion policy (ours), Action Chunking Transformer (ACT), fully connected network regression ({FCR}), and auto-encoder-based policy (AE+{FCR}).

We test all these methods on two datasets: one collected by our full automatic data extraction pipeline (see Sec.~\ref{subsec:data_extraction}) and a smaller, semi-automatic dataset where \zw{we manually discard outlier trajectories}. Therefore, the semi-automatic dataset is smaller but of higher quality. For both datasets, we measure the following metrics:
\begin{itemize}
\item \textbf{RMSE}: Average root mean square error between the predicted and the ground truth trajectory (meters).
\item \textbf{Dynamic Time Warping (DTW) distance}: The cumulative DTW distance between predicted and the ground truth trajectory and trajectory (meters).
\item \textbf{Iterative Closest Point (ICP) distance}: The mean distance between the predicted and the ground truth trajectory after aligning them using ICP (meters).
\item \textbf{Jerk}: Root mean square jerk the robot would experience if following the predicted trajectories ($m/s^{3}$).
\end{itemize}

For all metrics, a lower score indicates better performance, as shown in Table~\ref{tab:imitation_res}. The ``conditioning mode" specifies how the model was conditioned: ``Pre 2D" refers to conditioning on prior ball observations in image space, while ``Post 2D" refers to conditioning on future ball observations. Additionally, the ``action mode" specifies whether the policy outputs results in image space ($I_{\text{space}}$), which are then converted to Cartesian task space, or directly in task space ($T_{\text{space}}$).

We observe from Table \ref{tab:imitation_res} that conditioning on the future ball trajectory yields better predictions across most metrics compared to conditioning on the prior observed trajectory,
due to the additional future information guiding the wheelchair's movement.
The diffusion model outperforms ACT, {FCR}, and AE+{FCR} in RMSE by 20.6\%, 5.4\%, and 47.8\%, and in jerk by 71.84\%, 69.58\%, and 70.46\%, respectively. The diffusion model employs a stochastic denoising process to gradually refine the noise into a trajectory, which outperforms the traditional encoder-decoder structures that model the data in a limited embedding space \cite{vivekananthan2024comparativeanalysisgenerativemodels, chi2023diffusionpolicy, dhariwal2021diffusionmodelsbeatgans}.
Figure \ref{fig:diff_path} shows the diffusion trajectories alongside the ground truth \zw{(GT)} trajectories. Both quantitative and visual results demonstrate that the diffusion model can generate smoother trajectories that closely follow expert paths \zw{with small diffusion steps (10 steps) and inference time ($\approx0.1s$)}, \zw{which reveals the possibility to learn the motion planning and close the decision loop in the image space.}

\zw{
We conduct an additional ablation where the diffusion policy is conditioned on the predicted 2D ball trajectory but outputs the wheelchair trajectory directly in Cartesian space. This performs significantly worse than outputting in 2D space, validating the advantage of keeping the whole imitation process in image space.
}Additionally, we find that the trajectory prediction results from the automatic dataset are slightly worse than those from the semi-automatic datasets even the automatic dataset contains double size of data, which meets our expectations since the semi-automatic dataset has higher quality. Consequently, in subsequent evaluations, we opted to use diffusion policies with action predictions in image space trained from semi-automatic dataset.

\begin{figure*}[ht]
\centering
\begin{minipage}{0.38\textwidth}  
    \centering
    \captionof{table}{Comparison of success rate (SR) of different methods in hybrid(H) and real-court(R) settings.}
    \label{tab:harware_in_loop_results}
\resizebox{\textwidth}{!}{
\begin{tabular}{llcc}
\toprule
\textbf{Method} & \textbf{Conditioning} & \textbf{SR (H)} & \textbf{SR (R)} \\ \midrule
\makecell[l]{Diffusion (ours)} & \makecell[l]{Post 2D} & \textbf{97.67\%} & \textbf{68.49\%} \\ \midrule
\makecell[l]{{FCR}} & \makecell[l]{Post 2D} & \textbf{97.67\%} & 55.58\% \\ \midrule
TEB & \makecell[l]{Post 3D} & 37.21\% & 35.36\% \\ \bottomrule
\end{tabular}
}

    \vspace{0.05\textwidth} 

    \includegraphics[width=1.05\textwidth]{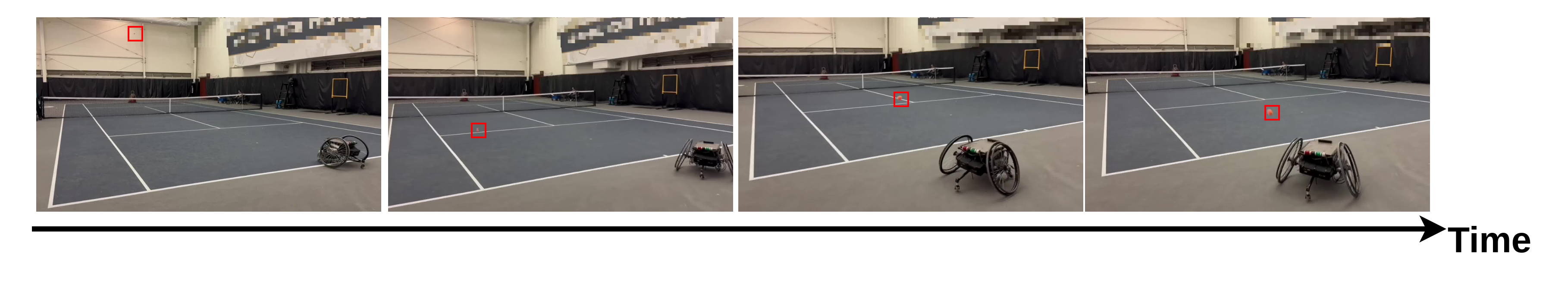}
    \captionof{figure}{Tennis robot responds to a ball launch over time. The ball is highlighted in the images.}
    \label{fig:realplay sequences}
\end{minipage}%
\hspace{0.009\textwidth}  
\begin{minipage}{0.26\textwidth}  
    \centering
    \includegraphics[width=\textwidth, height=2.05in]{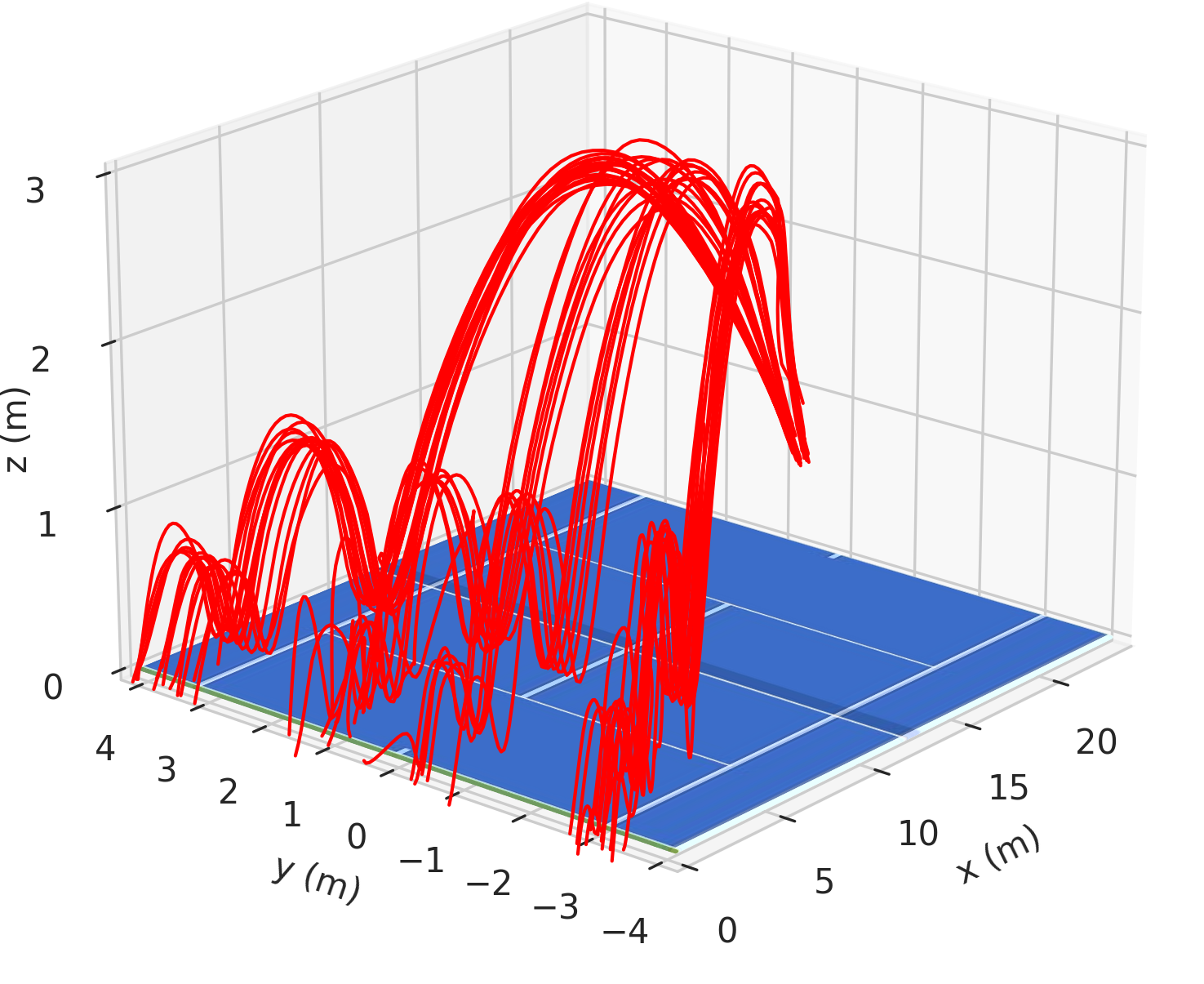}
    \captionof{figure}{Ball trajectory distribution of collected dataset.}
    \label{fig:distribution}
\end{minipage}
\hspace{0.002\textwidth}
\begin{minipage}{0.32\textwidth}  
    \centering
    \includegraphics[width=\textwidth, height=1.875in]{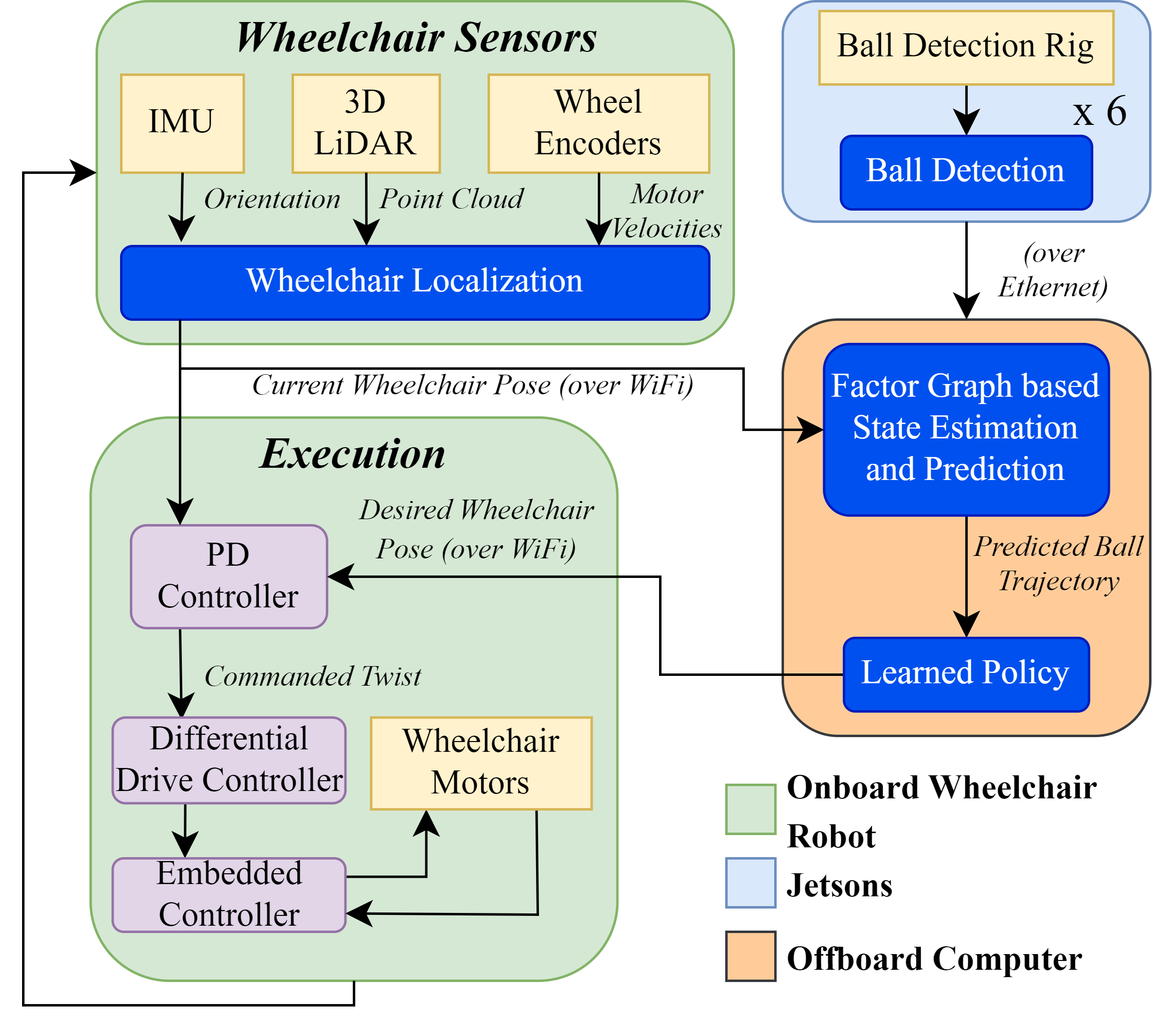}
    \captionof{figure}{Data flow between different components of the system for real-time experiment on a full-sized tennis court.}
    \label{fig:dataflow}
\end{minipage}
\vspace{-0.2in}
\end{figure*}

\subsection{Hardware-in-Loop Task Space Hybrid Test}
\label{subsec:task_space_evaluation}
\zz{
We conducted a series of hybrid tests, which combine real-world and simulated components, using a physical wheelchair robot and replayed real tennis ball trajectories collected on a tennis court. Balls were launched to various parts of the court \zw{(see Fig. \ref{fig:distribution} for trajectory distribution)}. These recorded trajectories were replayed, and the response of the physical wheelchair robot was recorded.
}

\zw{Since no expert ground truth trajectory is available in the task space, 
we use success rate as the evaluation metric. Success is defined as the wheelchair getting close enough for interception (within 1.4 meters of the ball) before it bounces three times (as wheelchair tennis permits up to two bounces).} We set the 1.4 m threshold based on previous wheelchair tennis research, which demonstrated that a robotic manipulator could successfully hit the ball when it was within this range~\cite{zaidi2023athletic}.
The results obtained are documented in Table~\ref{tab:harware_in_loop_results}. To compare performance, we also evaluated the best-performing imitation policies against a non-learning-based strategy from \cite{zaidi2023athletic}, referred to as the TEB baseline. The TEB baseline continuously monitors the ball’s predicted trajectory and identifies positions within the robot’s geometric hitting constraints. The Timed Elastic Band planner \cite{rosmann2017integrated}, a standard 2D navigation planner in ROS, estimates which point the robot can reach fastest, and then navigates to it.
\zw{
Both diffusion and FCR imitation policies trained with our framework outperform the TEB baseline, achieving a high success rate of 97.67\%.
}

\subsection{Real-time Experiment on a Tennis Court}
\label{subsec:real_world}
{
In this section, we present the results of our real-world, real-time experiment \zw{as an end-to-end evaluation} using a robot on a full-sized tennis court with real tennis balls.}

{\textbf{Setup:} 
\zz{Our system includes a wheelchair robot equipped with LiDAR, IMU, and wheel encoders, using an EKF for localization \cite{moore2016generalized}. This localization informs our policies and provides feedback to the Proportional-Derivative (PD) controller. Six cameras are placed around the court for tennis ball tracking, each connected to a Jetson Xavier for ball detection. The detections are transmitted via Ethernet to an offboard computer, where the predicted ball trajectory is computed for the imitation policy. To enable fast and asynchronous trajectory prediction of the tennis ball, we construct a factor graph \cite{xiao2024multi} to estimate the dynamic states of the ball, including its position, velocity, and spin. These estimated states serve as initial values for rolling out future predictions through an ordinary differential equation (ODE) solver. 
The data flow between components is shown in Fig.~\ref{fig:dataflow}.}


The results are tabulated in Table \ref{tab:harware_in_loop_results}, where it shows the diffusion policy \zw{with our learning framework outperforms the FCR and TEB by 12.91\% and 33.13\% in the real-world setting.} We provide snapshots of the policy execution on the physical robot in real-time with a real ball on a full-sized tennis court as illustrated in Fig.~\ref{fig:realplay sequences}.
The performance gap between experiments using real-time trajectories and those playing back recorded trajectories indicates that \zw{on-the-fly trajectory estimation in the large court has negative effects on the agile navigation task due to observation uncertainty and communication delays. The result shows diffusion-based motion planner is more robust to such system disturbance. }

\section{Conclusion and future work}

We propose a \zw{zero-shot knowledge transferring} framework that can transfer the wheelchair navigation skills from expert videos to our courts. By learning from web videos, our approach eliminates the need for static-goal, ego-centric, multi-view, or physical demonstrations, outperforming baselines in simulation and real-world experiments. In the future, we will mount a robotic manipulator on the wheelchair base and develop whole-body decision and control strategies that enable not only intercepting the ball but also successfully returning it. 
This work also opens new avenues for the research in human-robot teaming in the mixed adversarial-collaboration environment (Fig. \ref{fig:generalization}) where we also model the intention and behaviors of the environment agents \cite{ye2023learningadv, wu2023Adversarial}. 

\begin{figure}[ht]
    \centering    
    \includegraphics[width=0.8\linewidth]{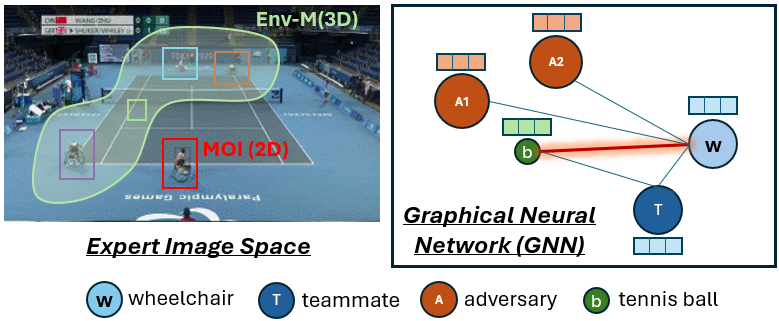}
    \caption{Our framework can generalize to multi-agent settings and various imitation policies. The plot depicts collaborative navigation learning with graph neural networks (GNNs).}
    \label{fig:generalization}
    \vspace{-5mm}
\end{figure}

\balance
\bibliographystyle{IEEEtran}
\bibliography{References}

\end{document}